\pdfoutput=1
\documentclass[a4paper,fleqn]{cas-sc}
\usepackage[square,comma,sort&compress,numbers]{natbib}
\usepackage{booktabs}
\usepackage{algorithm}
\usepackage{algorithmicx}
\usepackage{algpseudocode}
\def\tsc#1{\csdef{#1}{\textsc{\lowercase{#1}}\xspace}}
\tsc{WGM}
\tsc{QE}

\begin{document}
\let\WriteBookmarks\relax
\def\floatpagepagefraction{1}
\def\textpagefraction{.001}
\let\printorcid\relax

\shorttitle{A new fuzzy multi-attribute group decision-making method based on TOPSIS and optimization models}    

\shortauthors{Qixiao Hu et al.}

\title [mode = title]{A new fuzzy multi-attribute group decision-making method based on TOPSIS and optimization models}  

\tnotemark[0] 

\author[1]{Qixiao Hu}[type=editor,
    role=Researcher, 
]

\credit{Conceptualization of this study, Methodology, Software}

\affiliation[1]{organization={School of Mathematics, Sichuan University},
            addressline={610065}, 
            city={Chengdu},
            state={Sichuan},
            country={China}}

\author[1]{Shiquan Zhang}
\cormark[1]
\ead{shiquanzhang@scu.edu.cn}

\author[1]{Chaolang Hu}
\cormark[1]
\ead{huchaolang@scu.edu.cn}

\author[1]{Yuetong Liu}

\begin{abstract}
In this paper, a new method based on TOPSIS and optimization models is proposed for multi-attribute group decision-making in the environment of interval-valued intuitionistic fuzzy sets. Firstly, by minimizing the sum of differences between individual evaluations and the overall consistent evaluations of all experts, a new optimization model is established for determining expert weights. Secondly, based on TOPSIS method, the improved closeness index for evaluating each alternative is obtained. Finally, the attribute weight is determined by establishing an optimization model with the goal of maximizing the closeness of each alternative, and it is brought into the closeness index so that the alternatives can be ranked. Combining all these together, the complete fuzzy multi-attribute group decision-making algorithm is formulated, which can give full play to the advantages of subjective and objective weighting methods. In the end, the feasibility and effectiveness of the provided method are verified by a real case study.
\end{abstract}

\begin{highlights}
\item Unknown weights.
\item Optimization models that combine the advantages of objective and subjective methods.
\item Complete multi-attribute group decision-making method.
\end{highlights}


\maketitle

\section{Introduction}
Multi-attribute group decision-making is a process in which many decision-makers determine the decision-making range around the decision-making goal, and then propose decision-making methods to evaluate, rank and select alternatives \cite{1}. This process is mainly to solve the problems of evaluation and selection, and its theory and methods are widely used in engineering, technology, economy, management and other fields. In this paper, experts generally refer to decision-makers.

\par{Fuzzy multi-attribute group decision-making process can use many different methods, such as ELECTRE \cite{2}, PROMETHEE \cite{3}, TOPSIS \cite{4} and so on. However, no matter which method we use, we must consider how to deal with data fuzzily, expert weight and attribute weight. Firstly, with the development of economy and society, the decision-making problems that people need to solve are becoming more and more complicated. On the one hand, it is difficult to quantify some attributes because of their fuzziness. At this time, decision makers can't get accurate information, and accordingly, they can't make accurate evaluation. On the other hand, even if the attribute can be quantified, it is easy to make the evaluation value inaccurate due to the influence of subjective and objective factors such as the energy of decision makers and the incompleteness of understanding of things. It is not difficult to draw the conclusion that almost all the decision-making processes are related to fuzziness, which makes the problem of fuzzy multi-attribute decision-making aroused widespread concern. Fuzzy multi-attribute decision-making method introduces fuzzy theory into multi-attribute decision-making to improve the scientificity and practicability of decision-making, because it can not only better describe the attributes in alternatives, but also overcome the difficulty of inaccurate evaluation of decision makers caused by subjective and objective factors.}

\par{Secondly, both the determination of expert weight and attribute weight are very important in multi-attribute group decision-making, which has attracted the attention of a large number of scholars, because different weights may lead to different decision-making results. The methods of determining weights are divided into the following three categories: subjective weighting method, objective weighting method and combination of subjective and objective weighting method. The subjective weighting method is to compare the importance of decision makers or attributes and assign them. The advantage of this method is that it can be weighted according to the importance of decision makers or attributes, but it is subjective and adds a lot of manpower and material resources. Among them, the common methods are AHP \cite{12} and Delphi. The objective weighting method is to use objective data to obtain weights, which has the advantages of strong objectivity and strong mathematical theoretical basis. However, it does not consider the subjective intention of decision makers, and it will be inconsistent with the actual situation. Among them, the common methods are entropy weighting method \cite{13} and deviation maximization method. The combination of subjective and objective weighting method is aimed at the advantages and disadvantages of subjective weighting method and objective weighting method, which considers both the subjective intention of decision makers and the internal laws of objective data \cite{14}, thus making the results more real and reliable. Common methods include goal programming method. In practical application, we tend to use the combination of subjective and objective weighting method to determine the expert weight and attribute weight, which makes the final decision-making result more credible.}

\par{In order to solve the increasingly complex decision-making problems more reasonably, many scholars have conducted deeply research on fuzzy multi-attribute group decision-making methods. In the environment of intuitionistic fuzzy set, Sina et al. directly gives the expert weight by subjective weighting method, and determines the attribute weight by combining CRITIC and Ideal Point that are objective weighting method, then it gives the alterative ranking by combining ARAS and EDAS to solve the decision-making problem of entrepreneur construction projects \cite{15}. In the environment of interval-valued intuitionistic fuzzy sets, Ting-Yu directly gives the expert weight by subjective weighting method, and determines the attribute weight by weight optimization model, finally it gives the alterative ranking by TOPSIS method to solve the treatment plan decision-making problem \cite{16}. In the environment of intuitionistic fuzzy set,  Shi-fang et al. obtains the expert weights through IFWA operator and aggregates the decision matrices corresponding to multiple experts into a decision matrix, and determines the attribute weights through intuitionistic fuzzy entropy, then finally solves the personnel decision-making problem through GRA \cite{17}. In the environment of intuitionistic fuzzy set, Behnam et al. obtains the expert weight through IFWA operator, then determines the attribute weight through subjective weighting method combined with IFWA operator, and solves the decision-making problem of company updating manufacturing system by combining ELECTRE method \cite{18}. In the environment of interval-valued intuitionistic fuzzy sets, Feifei et al. directly gives the expert weight by subjective weighting method, then determines the attribute weight by continuous weighted entropy, finally solves the evaluation problem of community emergency risk management by TOPSIS \cite{19}. In the environment of interval hesitant fuzzy sets, Gitinavard et al. determines the attribute weight by combining expert empowerment with extended maximum deviation method, and extends IVHF-TOPSIS method to determine the expert weight, then uses the proposed IVHF-MCWR model to solve the location and supplier decision-making problems \cite{20}.}

\par{When fuzzy multi-attribute group decision-making is used to solve the above problems, the expert weight or attribute weight is determined by subjective weighting method or objective weighting method, which cannot take into account the internal laws of data itself and expert opinions at the same time. Liu et al. put forward a new model of expert weight optimization \cite{21}. When the expert evaluation results are consistent with those of all expert groups, we should give him higher weight. At the same time, we can let decision makers give constraints on the expert weight, which full develops the advantages of subjective and objective weighting methods. However, Ting-Yu directly entrusts the weight of experts \cite{16}, thus this paper extends the optimization model of determining the weight of experts proposed by \cite{21} under the environment of interval-valued intuitionistic fuzzy sets and combines it with \cite{16}. So that the weight of experts and attributes will be determined by the optimization model formed by objective data respectively, and at the same time, it can be constrained by expert opinions. Finally, a complete fuzzy multi-attribute group decision-making process is formed by combining TOPSIS.}

\par{This paper is arranged as follows. In the second part, the related theories of interval intuitionistic fuzzy sets are expounded. The third part explains the extended expert weight determination method \cite{21} and how to determine the attribute weight, as well as the extended TOPSIS \cite{21}, and finally develops a complete fuzzy multi-attribute group decision-making method. The fourth part illustrates the effectiveness of this method through a decision-making case, and the fifth part is the summary of this paper.}

\section{Preliminaries}
\newtheorem{definition}{Def}
Fuzzy set is a common method for fuzzy processing. In 1965, Zedah put forward the concept of fuzzy set \cite{5}, which provided a solution for people to deal with fuzzy information in decision-making problems.In 1986, Atanassov et al. extended the fuzzy set and put forward the intuitionistic fuzzy set \cite{6}. This theory can simultaneously express the support, opposition and neutrality of the decision-maker in terms of membership, non-membership and hesitation, which can effectively deal with the problem of uncertain decision information. In 1996, Gehrke et al. advanced interval fuzzy sets to solve the problem that it is too strict to use a certain numerical value as the membership degree \cite{7}, and its membership degree is in the form of a closed subinterval of an interval. In 2009, Torra et al. proposed the concept of hesitant fuzzy set in order to describe the hesitation in the decision-making process \cite{8}, which allows the existence of multiple membership values. In 2012, Zhu et al. combined hesitant fuzzy sets with intuitive fuzzy sets, and proposed dual hesitant fuzzy sets \cite{9}. They added non-membership degree to the hesitant fuzzy set, and allowed many values to appear in the non-membership degree. In 2013, Yager proposed Pythagorean fuzzy sets by adjusting the constraints of membership and non-membership in intuitionistic fuzzy sets \cite{10}. Other types of fuzzy sets, such as interval intuitionistic fuzzy sets \cite{11}, are all extended on the basis of the above fuzzy sets.

\par{The multi-attribute group decision-making algorithm proposed in this paper is discussed in the environment of interval intuitionistic fuzzy sets, so we will list the related theories of interval intuitionistic fuzzy sets.}

\begin{definition}
$X$ be a non-empty set and the interval intuitionistic fuzzy set is as follows:
\begin{equation}
   A=\left \{<x,(\mu_{A}(x),v_{A}(x))>|x\in X \right\}.
\end{equation}
Where $\mu_{A}(x)$ and $v_{A}(x)$ represent the membership interval and non-membership interval of $x\in X$, respectively. They can be expressed by the interval as:
\begin{equation}
\mu_{A}(x)=[\mu_{A}^{-}(x),\mu_{A}^{+}(x)],\quad v_{A}(x)=[v_{A}^{-}(x),v_{A}^{+}(x)],
\end{equation}
they satisfy: $\mu_{A}(x)\subseteq [0,1]$, $v_{A}(x)\subseteq [0,1]$ and  $0\le \mu_{A}(x)+v_{A}(x)\le 1$. When $\mu_{A}^{-}(x)=\mu_{A}^{+}(x)$ and $v_{A}^{-}(x)=v_{A}^{+}(x)$ , interval intuitionistic fuzzy sets degenerate into intuitionistic fuzzy sets. At the same time, for $\forall x\in X$ , its hesitation interval can be expressed as:
\begin{equation}
\pi_{A}(x)=[\pi_{A}^{-}(x),\pi_{A}^{+}(x)]=[1-\mu_{A}^{+}(x)-v_{A}^{+}(x),1-\mu_{A}^{-}(x)-v_{A}^{-}(x)].
\end{equation}
\end{definition}

\begin{definition}
If
\begin{equation}
\begin{split}
A_{x}=<\mu_{A}(x),v_{A}(x)>=<[\mu_{A}^{-}(x),\mu_{A}^{+}(x)], [v_{A}^{-}(x),v_{A}^{+}(x)]>,\\
B_{x}=<\mu_{B}(x),v_{B}(x)>=<[\mu_{B}^{-}(x),\mu_{B}^{+}(x)], [v_{B}^{-}(x),v_{B}^{+}(x)]>,\nonumber
\end{split}
\end{equation}
are any two interval intuitionistic fuzzy sets, $\lambda$ is any real number greater than 0, then there are the following operation rules:
\begin{enumerate}
\item $A_{x}\oplus B_{x}=<[\mu_{A}^{-}(x)+\mu_{B}^{-}(x)-\mu_{A}^{-}(x)\cdot \mu_{B}^{-}(x), \mu_{A}^{+}(x)+\mu_{B}^{+}(x)-\mu_{A}^{+}(x)\cdot \mu_{B}^{+}(x)], [v_{A}^{-}(x)\cdot v_{B}^{-}(x), v_{A}^{+}(x)\cdot v_{B}^{+}(x)]>;$
\item $A_{x}\otimes B_{x}=<[\mu_{A}^{-}(x)\cdot \mu_{B}^{-}(x), \mu_{A}^{+}(x)\cdot \mu_{B}^{+}(x)], [v_{A}^{-}(x)+v_{B}^{-}(x)-v_{A}^{-}(x)\cdot v_{B}^{-}(x), v_{A}^{+}(x)+v_{B}^{+}(x)-v_{A}^{+}(x)\cdot v_{B}^{+}(x)]>;$
\item $\lambda\cdot A_{x}=<[1-(1-\mu_{A}^{-}(x))^{\lambda}, 1-(1-\mu_{A}^{+}(x))^{\lambda}], [(v_{A}^{-}(x))^{\lambda}, (v_{A}^{+}(x))^{\lambda}]>;$
\item $(A_{x})^{\lambda}=<[(\mu_{A}^{-}(x))^{\lambda}, (\mu_{A}^{+}(x))^{\lambda}], [1-(1-v_{A}^{-}(x))^{\lambda}, 1-(1-v_{A}^{+}(x))^{\lambda}]>.$
\end{enumerate} 
\end{definition}

\begin{definition}
If
\begin{equation}
\begin{split}
A_{x}=<\mu_{A}(x),v_{A}(x)>=<[\mu_{A}^{-}(x),\mu_{A}^{+}(x)], [v_{A}^{-}(x),v_{A}^{+}(x)]>,\\
B_{x}=<\mu_{B}(x),v_{B}(x)>=<[\mu_{B}^{-}(x),\mu_{B}^{+}(x)], [v_{B}^{-}(x),v_{B}^{+}(x)]>,\nonumber
\end{split}
\end{equation}
are any two interval intuitionistic fuzzy sets, then the lower bound $p^{-}(A_{x}\supseteq B_{x})$ of the inclusion comparison possibility of $A_{x}$ and $B_{x}$ is defined as \cite{22}:
\begin{equation}
p^{-}(A_{x}\supseteq B_{x})=max\left\{1-max\left\{\frac{(1-v_{B}^{-}(x))-\mu_{A}^{-}(x)}{(1-\mu_{A}^{-}(x)-v_{A}^{+}(x))+(1-\mu_{B}^{+}(x)-v_{B}^{-}(x))}, 0\right\}, 0\right\}.
\end{equation}
And the upper bound $p^{+}(A_{x}\supseteq B_{x})$ of the inclusion comparison possibility of $A_{x}$ and $B_{x}$ is defined as:
\begin{equation}
p^{+}(A_{x}\supseteq B_{x})=max\left\{1-max\left\{\frac{(1-v_{B}^{+}(x))-\mu_{A}^{+}(x)}{(1-\mu_{A}^{+}(x)-v_{A}^{-}(x))+(1-\mu_{B}^{-}(x)-v_{B}^{+}(x))}, 0\right\}, 0\right\}.
\end{equation}
Then the inclusion comparison possibility $p(A_{x}\supseteq B_{x})$ of $A_{x}$ and $B_{x}$ is defined as:
\begin{equation}
p(A_{x}\supseteq B_{x})=\frac{1}{2}(p^{-}(A_{x}\supseteq B_{x})+p^{+}(A_{x}\supseteq B_{x})),
\end{equation}
that is to say, the possibility that $A_{x}$ is not smaller than $B_{x}$ is $p(A_{x}\supseteq B_{x})$. Then $p(A_{x}\supseteq B_{x})$ has the following properties:
\begin{enumerate}
\item $0\le p(A_{x}\supseteq B_{x})\le 1;$
\item $p(A_{x}\supseteq B_{x})+p(A_{x}\subseteq B_{x})=1.$
\end{enumerate} 
\end{definition}

\section{Improved fuzzy multi-attribute group decision-making method}

\subsection{Fuzzy multi-attribute group decision-making problem}
With the increasing complexity of decision-making environment and problems, fuzzy multi-attribute group decision-making methods for solving evaluation and decision-making problems have been widely concerned. Although there are many methods for fuzzy multi-attribute group decision-making, we all have to go through three steps: fuzzification, expert weight and attribute weight determination. When solving the multi-attribute group decision-making problem, we might as well make the following assumptions:
\begin{enumerate}
\item $l$ decision makers;
\item $m$ alternatives;
\item $n$ indicators of each alternative.
\end{enumerate}
Experts need to evaluate each index of each alternative semantically, among which the $k-$th decision maker is marked as $D_{k}$, the $i-$th alternative is marked as $A_{i}$, and the $j-$th indicator is marked as $x_{j}$. The following Figure \ref{figure2} shows the process of solving the fuzzy multi-attribute group decision-making problem.
\begin{figure}[pos=h]
\centering
\includegraphics[height=3.84cm,width=16cm]{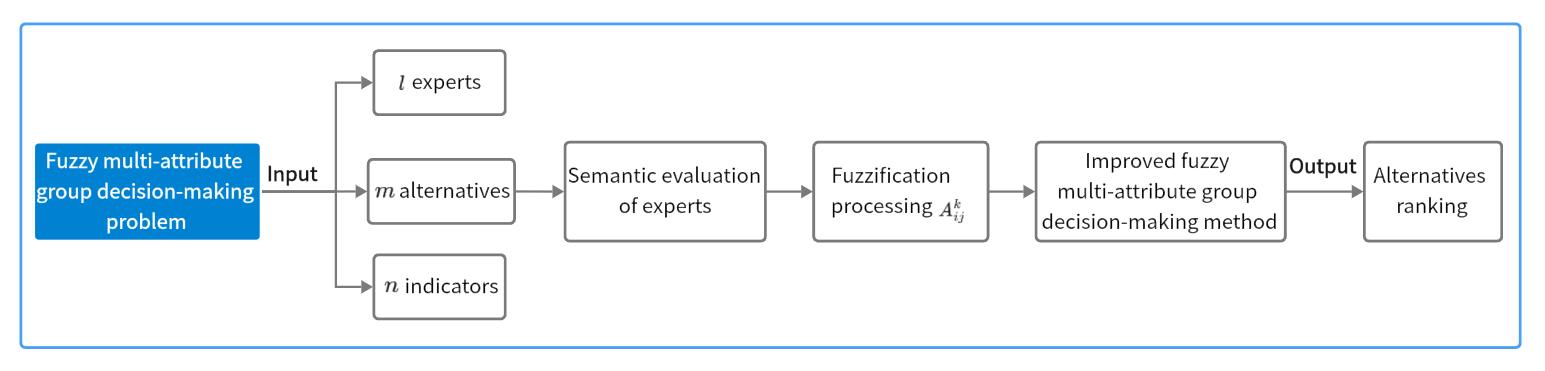}
\caption{Fuzzy multi-attribute group decision-making problem}
\label{figure2}
\end{figure}

\par{Specifically, based on the improved TOPSIS method proposed in \cite{16}, this paper first fuzzifies the semantic evaluation of the $j-$th index of the $i-$th alternative by the $k-$th decision-maker through the interval intuitionistic fuzzy set, which can be recorded as: $A_{ij}^{k}=<[\mu_{ij}^{k-}, \mu_{ij}^{k+}], [v_{ij}^{k-}, v_{ij}^{k+}]>$. Secondly, optimization models are established to determine the expert weight and attribute weight, and corresponding constraints can be added according to actual needs. In this way, we can not only give full play to the advantages of objective weighting method that make full use of objective data, but also avoid the disadvantages of not considering the subjective intention of decision makers. Finally, the alternatives can be sorted. Thus, a complete fuzzy multi-attribute group decision-making method is formed, which can be used to help people solve complex multi-attribute group decision-making problems.}

\subsection{Determination of expert weight based on optimization model}
Liu et al. points out that different experts have different degrees of experience and knowledge of related fields, therefore the importance of different experts should be different, and we should give higher weight to experts with rich experience and full understanding of decision-making projects \cite{21}. In other words, if the evaluation results of an expert are more consistent with the evaluation results of all experts, the evaluation results of the expert will be more valuable for reference, so we give such experts greater weight. Then we will establish an optimization model based on this and combine the subjective intention of decision makers to restrict it, so that we can get the weight of experts through the combination of subjective and objective methods. Firstly, we assume that the decision matrix corresponding to the $i-$th alternative is $A_{(i)}$:
\begin{equation}
A_{(i)}=\left(\begin{array}{cccc} A_{(i)}^{1} &  A_{(i)}^{2} & \cdots &  A_{(i)}^{l}\end{array}\right),
\end{equation}
where $A_{(i)}^{k}=\left(\begin{array}{cccc} A_{i1}^{k} &  A_{i2}^{k} & \cdots &  A_{in}^{k}\end{array}\right)^{T}$ represents the evaluation of the $i-$th alternative by the $k-$th expert, and we assume that the corresponding weight of experts is $\boldsymbol{w}=\left(\begin{array}{cccc} w_{1} &  w_{2} & \cdots &  w_{l}\end{array}\right)^{T}$. Secondly, each alternative corresponds to a consistent score point, which is obtained by linear combination of $l$ experts' evaluations. The interval intuitionistic fuzzy set corresponding to the evaluation of the $j-$th indicator of the $i-$th alternative by the $k-$th decision-maker is $A_{ij}^{k}$, which corresponds to four numbers. In order to be able to use the weight determination model proposed by \cite{21}, we split the interval intuitionistic fuzzy set, thus the length of the evaluation column vector of the $k-$th decision-maker for the $i-$th alternative becomes four times as long as the original one:
\begin{equation}
A_{(i)}^{k}=\left(\begin{array}{cccccccccccc} \mu_{i1}^{k-} & \cdots & \mu_{in}^{k-} & \mu_{i1}^{k+} & \cdots & \mu_{in}^{k+} & v_{i1}^{k-} & \cdots & v_{in}^{k-} & v_{i1}^{k+} & \cdots & v_{in}^{k+}\end{array}\right)^{T}.
\end{equation}
Then the consistent score point corresponding to the $i-$th alternative can be expressed as: 
\begin{equation}\label{b}
\boldsymbol{b}_{(i)}=\sum_{k=1}^{l} w_{k}\cdot A_{(i)}^{k}
=\left(\begin{array}{cccccccccccc}
\sum_{k=1}^{l} w_{k}\cdot\mu_{i1}^{k-}\\
\vdots\\
\sum_{k=1}^{l} w_{k}\cdot\mu_{in}^{k-}\\
 \sum_{k=1}^{l} w_{k}\cdot\mu_{i1}^{k+}\\
 \vdots\\
 \sum_{k=1}^{l} w_{k}\cdot\mu_{in}^{k+}\\
 \sum_{k=1}^{l} w_{k}\cdot v_{i1}^{k-}\\
 \vdots\\
 \sum_{k=1}^{l} w_{k}\cdot v_{in}^{k-}\\
 \sum_{k=1}^{l} w_{k}\cdot v_{i1}^{k+}\\
 \vdots\\
\sum_{k=1}^{l} w_{k}\cdot v_{in}^{k+}
\end{array}\right),
\end{equation}
which reflects the overall evaluation results of experts on the $i-$th alternative. 

\par{And all alternatives are treated equally, so the decision matrix $A_{(i)}$ corresponding to each alternative can be assembled into an overall decision matrix $A=\left(\begin{array}{cccc} A_{(1)} &  A_{(2)} & \cdots &  A_{(m)}\end{array}\right)^{T}$ when determining the expert weight, where the $k-$th column $A^{(k)}$ of $A$ represents the evaluation results of all alternatives by the $k-$th expert. Then the overall consistent score point is $\boldsymbol{b}=\left(\begin{array}{cccc} \boldsymbol{b}_{(1)} &  \boldsymbol{b}_{(2)} & \cdots &  \boldsymbol{b}_{(m)}\end{array}\right)^{T}$, thus the distance from the evaluation results of all alternatives by the $k-$th expert to the overall consistent score point $\boldsymbol{b}$ is:
\begin{equation}
d_{(k)}=\left \| A^{(k)}-\boldsymbol{b} \right \|_{2}.
\end{equation}}

\par{Finally, we can establish an optimization model as follows:
\begin{equation}
\begin{split}
&\min_{\boldsymbol{w}} Q(\boldsymbol{w})=\sum_{k=1}^{l}d_{(k)}\\
&s.t.\begin{cases}\sum_{k=1}^{l}w_{k}=1,\\
0\le w_{k}\le 1, \quad 1\le k \le l.\label{eq1}
\end{cases}
\end{split}
\end{equation}
Through the optimization model, we can understand that if the distance between the evaluation result of an expert and the overall consistent score is closer, we will give such an expert greater weight. For example, as shown in Figure \ref{figure}, after the above-mentioned optimization model processing, the obtained expert weights are ranked as $w_{4}>w_{2}>w_{1}>w_{5}>w_{3}$. At the same time, Liu et al. also proves that the optimization model has a unique solution \cite{21}, and we can attach more constraints to the optimization model, such as giving the highest weight to authoritative experts, and the optimization model still has a unique solution. And numerical experiments show that the closer the expert's evaluation result is to the consistent score point, the higher the weight he will get.
\begin{figure}[pos=h]
\centering
\includegraphics[height=4.7cm,width=5.5cm]{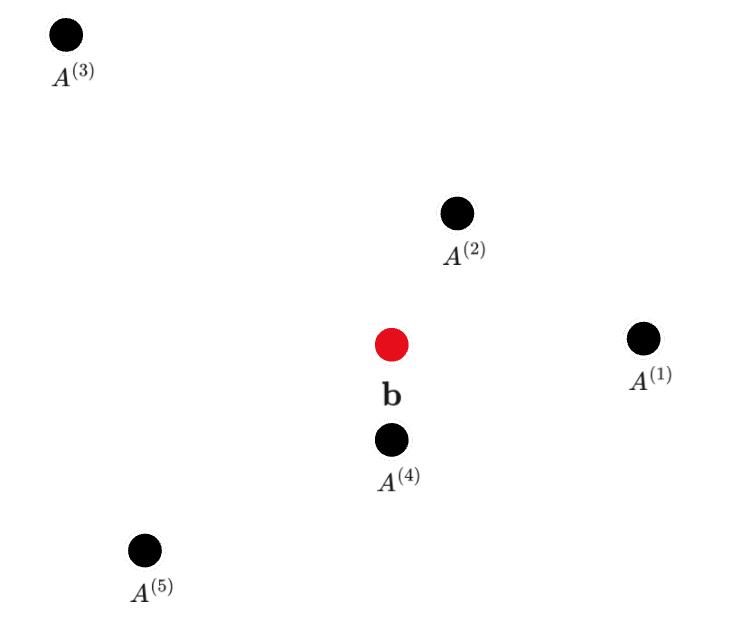}
\caption{Relationship between weight and distance}
\label{figure}
\end{figure}}

\subsection{Improved TOPSIS method}
An improved TOPSIS method is proposed to solve the problem of multi-attribute group decision-making by \cite{16}. In this paper, the weight of experts is given directly, but its acquisition method is not explained. In the last section, we can calculate the expert weight by extending and using the method proposed by \cite{21}. Therefore, after getting the expert weight, we can solve the multi-attribute group decision-making problem completely by combining with the improved TOPSIS proposed by \cite{16}. The specific process is as follows.

\par{
Firstly, after obtaining the expert weight $\boldsymbol{w}$ according to the optimization model, we can weight the evaluation $A_{ij}^{k}$ of the $j-$th indicator of the $i-$th alternative by the $k-$th decision-maker to get ${\mathop{A}\limits _{\cdot}}_{ij}^{k}$:
\begin{equation}\label{eq3}
{\mathop{A}\limits _{\cdot}}_{ij}^{k}=l\cdot w_{k}\cdot A_{ij}^{k}\triangleq <[{\mathop{\mu}\limits_{\cdot}}_{ij}^{k-}, {\mathop{\mu}\limits_{\cdot}}_{ij}^{k+}], [{\mathop{v}\limits_{\cdot}}_{ij}^{k-}, {\mathop{v}\limits_{\cdot}}_{ij}^{k+}]>.
\end{equation}
Secondly, according to \cite{16}, the optimal membership degree $p({\mathop{A}\limits _{\cdot}}_{ij}^{k})$ of ${\mathop{A}\limits _{\cdot}}_{ij}^{k}$ can be calculated according to the following formula:
\begin{equation}\label{eq4}
p({\mathop{A}\limits _{\cdot}}_{ij}^{k})=\frac{1}{l(l-1)}(\sum_{k^{\prime}=1}^{l}p({\mathop{A}\limits _{\cdot}}_{ij}^{k}\supseteq {\mathop{A}\limits _{\cdot}}_{ij}^{k^{\prime}})+\frac{l}{2}-1).
\end{equation}
According to the above calculation, the interval intuitionistic fuzzy order weighted average (IIOWA) operator \cite{23,24} can be extended to get the comprehensive decision matrix $D$. The specific steps are as follows:
\begin{enumerate}
\item Reorder $\left(\begin{array}{cccc} 1 & 2 & \cdots & l \end{array}\right)$ to $\left(\begin{array}{cccc} \sigma(1) & \sigma(2) & \cdots & \sigma(l) \end{array}\right)$, which satisfies $p({\mathop{A}\limits _{\cdot}}_{ij}^{\sigma(k-1)}) \ge p({\mathop{A}\limits _{\cdot}}_{ij}^{\sigma(k)})$;
\item Calculate the weight vector $\boldsymbol{\tau}= \left(\begin{array}{cccc} \tau_{1} & \tau_{2} & \cdots & \tau_{l} \end{array}\right)^{T}$ of IIOWA operator:
\begin{equation}
\tau_{k}=\frac{e^{-((k-u_{l})^{2}/2\cdot t_{l}^{2})}}{\sum_{k^{\prime}=1}^{l}e^{-((k^{\prime}-u_{l})^{2}/2\cdot t_{l}^{2})}},
\end{equation}
where $u_{l}$ is the average value of $1, 2, \cdots , l$ and $t_{l}$ is the corresponding standard deviation.
\item Using IIOWA operator to calculate the element $A_{ij}$ in row $i$ and column $j$ of the comprehensive decision matrix $D$:
\begin{equation}\nonumber
A_{ij} = <[1-\prod_{k=1}^{1}(1-{\mathop{\mu}\limits _{\cdot}}_{ij}^{\sigma(k)-})^{\tau_{k}} , 1-\prod_{k=1}^{1}{\mathop{\mu}\limits _{\cdot}}_{ij}^{\sigma(k)+})^{\tau_{k}}], [\prod_{k=1}^{1}({\mathop{v}\limits _{\cdot}}_{ij}^{\sigma(k)-})^{\tau_{k}}, \prod_{k=1}^{1}({\mathop{v}\limits _{\cdot}}_{ij}^{\sigma(k)+})^{\tau_{k}}]>,
\end{equation}
which represents the comprehensive evaluation of the $j-$th indicator of the $i-$th alternative by all experts, and is abbreviated as $A_{ij}=<[\mu_{ij}^{-}, \mu_{ij}^{+}], [v_{ij}^{-}, v_{ij}^{+}]>$.
\end{enumerate}
}

\par{
The comprehensive decision matrix $D$ obtained by the above two steps can not only reflect the importance of different experts, but also reflect the consistency of all experts' evaluation. Then, according to the comprehensive decision matrix $D$, the positive and negative ideal solutions of interval intuitionistic fuzzy are found:
\begin{align}
A_{+}=\left\{ <x_{j}, ([\mu_{+j}^{-}, \mu_{+j}^{+}], [v_{+j}^{-}, v_{+j}^{+}])> | x_{j}\in X, j=1,2,\cdots, n\right\},\label{eq5.1}\\
A_{-}=\left\{ <x_{j}, ([\mu_{-j}^{-}, \mu_{-j}^{+}], [v_{-j}^{-}, v_{-j}^{+}])> | x_{j}\in X, j=1,2,\cdots, n\right\},\label{eq5.2}
\end{align}
where
\begin{equation}\nonumber
\begin{split}
[\mu_{+j}^{-}, \mu_{+j}^{+}]=[((\max_{i}\mu_{ij}^{-}|x_{j}\in X_{b}), (\min_{i}\mu_{ij}^{-}|x_{j}\in X_{c})), ((\max_{i}\mu_{ij}^{+}|x_{j}\in X_{b}), (\min_{i}\mu_{ij}^{+}|x_{j}\in X_{c}))],\\
[v_{+j}^{-}, v_{+j}^{+}]=[((\min_{i}v_{ij}^{-}|x_{j}\in X_{b}), (\max_{i}v_{ij}^{-}|x_{j}\in X_{c})), ((\min_{i}v_{ij}^{+}|x_{j}\in X_{b}), (\max_{i}v_{ij}^{+}|x_{j}\in X_{c}))],\\
[\mu_{-j}^{-}, \mu_{-j}^{+}]=[((\min_{i}\mu_{ij}^{-}|x_{j}\in X_{b}), (\max_{i}\mu_{ij}^{-}|x_{j}\in X_{c})), ((\min_{i}\mu_{ij}^{+}|x_{j}\in X_{b}), (\max_{i}\mu_{ij}^{+}|x_{j}\in X_{c}))],\\
[v_{-j}^{-}, v_{-j}^{+}]=[((\max_{i}v_{ij}^{-}|x_{j}\in X_{b}), (\min_{i}v_{ij}^{-}|x_{j}\in X_{c})), ((\max_{i}v_{ij}^{+}|x_{j}\in X_{b}), (\min_{i}v_{ij}^{+}|x_{j}\in X_{c}))].
\end{split}
\end{equation}
Also, $X_{b}$ represents the benefit indicator, and $X_{c}$ represents the cost indicator in the indicator. Finally, assuming that the attribute weight is $\overline{\boldsymbol{w}}=\left(\begin{array}{cccc} {\overline{w}}_{1} & {\overline{w}}_{2} & \cdots & {\overline{w}}_{n} \end{array}\right)^{T}$, we can improve the closeness index of the $i-$th alternative to be $CC(A_{i})$:
\begin{equation}\label{eq6}
\begin{split}
CC(A_{i})=&\sum_{j=1}^{n}p((A_{ij}\supseteq A_{-j}|x_{j}\in X_{b}), (A_{-j}\supseteq A_{ij}|x_{j}\in X_{c}))\overline{w}_{j}\cdot \\
&\left\{\sum_{j=1}^{n}[(p(A_{+j}\supseteq A_{ij})+p(A_{ij}\supseteq A_{-j})|x_{j}\in X_{b}), (p(A_{ij}\supseteq A_{+j})+p(A_{-j}\supseteq A_{ij})|x_{j}\in X_{c})]\overline{w}_{j} \right\}^{-1},
\end{split}
\end{equation}
where $0\le CC(A_{i})\le 1 (i=1,2,\cdots, m)$. For the $j-$th indicator of the $i-$th alternative, if it belongs to the benefit indicator, the inclusion comparison possibility $p(A_{ij}\supseteq A_{-j})$ with $A_{ij}$ not less than $A_{-j}$ and the inclusion comparison possibility $p(A_{+j}\supseteq A_{ij})$ with $A_{ij}$ not greater than $A_{+j}$ are calculated. At this time, if there is a higher possibility that $A_{ij}$ is better than $A_{+j}$ and a lower possibility that $A_{ij}$ is worse than $A_{-j}$, then the $j-$th indicator of the $i-$th alternative has a good performance. And the same is true for the cost indicator. Therefore, we can sort the closeness $CC(A_{i})$ and choose the alternative with the largest index value as the optimal alternative.
}

\subsection{Determination method of attribute weight}
In the first two parts, the problem of multi-attribute group decision-making can be solved by combining the method of determining expert weights proposed 
by \cite{21} with the improved TOPSIS method in \cite{16}, but how to calculate attribute weights is not explained. When the attribute weight is unknown, Ting-Yu also suggests that an optimization model can be established by combining subjective and objective weighting methods to determine the attribute weight \cite{16}. The specific methods are as follows.

\par{
In the previous part, we finally choose the alternative through the improved closeness $CC(A_{i})$. When the attribute weights are unknown, Ting-Yu established the following optimization model \cite{16}:
\begin{equation}
\begin{split}
&\max\left\{ CC(A_{1}), CC(A_{2}), \cdots, CC(A_{m})\right\}\\
&s.t. \begin{cases}
\sum_{j=1}^{n}\overline{w}_{j}=1,\\ 
\overline{w}_{j}\ge 0, \quad j=1,2,\cdots, n.
\end{cases}
\end{split}
\end{equation}

At this time, the above multi-objective optimization model is transformed into the following single-objective optimization model by using the max-min operator proposed in \cite{25}:
\begin{equation}
\begin{split}
&\max \vartheta\\
&s.t.
\begin{cases}
CC(A_{i})\ge \vartheta,\quad i=1,2,\cdots, m,\\
\left(\begin{array}{cccc} {\overline{w}}_{1} & {\overline{w}}_{2} & \cdots & {\overline{w}}_{n} \end{array}\right)\in \Gamma _{0}.
\end{cases}
\end{split}
\end{equation}
Where
\begin{equation}
\Gamma _{0}=\left\{ \left(\begin{array}{cccc} {\overline{w}}_{1} & {\overline{w}}_{2} & \cdots & {\overline{w}}_{n} \end{array}\right)|\sum_{j=1}^{n}\overline{w}_{j}=1, \overline{w}_{j}\ge 0, j=1,2,\cdots, n\right\},
\end{equation}
and attribute weights can be obtained by solving the above optimization model.}

\par{In practical application, experts can limit the attribute weight according to their own experience and professional knowledge, which can be divided into five forms: weak ranking, strict ranking, ranking difference, interval boundary and proportional boundary. However, the opinions of experts are almost impossible to be completely unified, so the following non-negative deviation variables:
\begin{equation}
e_{(1)j_{1}j_{2}}^{-},\quad e_{(2)j_{1}j_{2}}^{-},\quad e_{(3)j_{1}j_{2}j_{3}}^{-},\quad e_{(4)j_{1}}^{-},\quad e_{(4)j_{1}}^{+},\quad e_{(5)j_{1}j_{2}}^{-}\quad (j_{1}\neq j_{2}\neq j_{3}),
\end{equation}
which can be added to the five types of constraints to become the relaxed five types of constraints.
\begin{enumerate}
\item Relaxed weak ranking:
\begin{equation}
\Gamma_{1}=\left\{\left(\begin{array}{cccc} {\overline{w}}_{1} & {\overline{w}}_{2} & \cdots & {\overline{w}}_{n} \end{array}\right)\in \Gamma _{0}|\overline{w}_{j_{1}}+e_{(1)j_{1}j_{2}}^{-}\ge \overline{w}_{j_{2}},j_{1}\in \Upsilon_{1}, j_{2}\in \Lambda _{1}\right\},\nonumber
\end{equation}
where $\Upsilon_{1}$ and $\Lambda _{1}$ are two disjoint subsets in index set $N=\left\{1,2,\cdots,n\right\}$.
\item Relaxed strict ranking:
\begin{equation}
\Gamma_{2}=\left\{\left(\begin{array}{cccc} {\overline{w}}_{1} & {\overline{w}}_{2} & \cdots & {\overline{w}}_{n} \end{array}\right)\in \Gamma _{0}|\overline{w}_{j_{1}}-\overline{w}_{j_{2}}+e_{(2)j_{1}j_{2}}^{-}\ge \delta_{j_{1}j_{2}}^{\prime},j_{1}\in \Upsilon_{2}, j_{2}\in \Lambda _{2}\right\},\nonumber
\end{equation}
where $\delta_{j_{1}j_{2}}^{\prime}$ is a constant and  $\delta_{j_{1}j_{2}}^{\prime}\ge 0$, $\Upsilon_{2}$ and $\Lambda _{2}$ are two disjoint subsets in index set $N$.
\item Relaxed ranking of differences:
\begin{equation}
\Gamma_{3}=\left\{\left(\begin{array}{cccc} {\overline{w}}_{1} & {\overline{w}}_{2} & \cdots & {\overline{w}}_{n} \end{array}\right)\in \Gamma _{0}|\overline{w}_{j_{1}}-2\overline{w}_{j_{2}}+\overline{w}_{j_{3}}+e_{(3)j_{1}j_{2}j_{3}}^{-}\ge 0,j_{1}\in \Upsilon_{3}, j_{2}\in \Lambda _{3}, j_{3}\in \Omega_{3}\right\},\nonumber
\end{equation}
where $\Upsilon_{3}$, $\Lambda _{3}$ and $\Omega_{3}$ are three disjoint subsets in index set $N$.
\item Relaxed interval boundary:
\begin{equation}
\Gamma_{4}=\left\{\left(\begin{array}{cccc} {\overline{w}}_{1} & {\overline{w}}_{2} & \cdots & {\overline{w}}_{n} \end{array}\right)\in \Gamma _{0}|\overline{w}_{j_{1}}+e_{(4)j_{1}}^{-}\ge \delta_{j_{1}}, \overline{w}_{j_{1}}-e_{(4)j_{1}}^{+}\le \delta_{j_{1}}+\varepsilon_{j_{1}}, j_{1}\in \Upsilon_{4}\right\},\nonumber
\end{equation}
where $\delta_{j_{1}}$ and $\varepsilon_{j_{1}}$ are constans, which satisfy $\delta_{j_{1}}\ge 0$, $\varepsilon_{j_{1}}\ge 0$, $0\le \delta_{j_{1}}\le \delta_{j_{1}}+\varepsilon_{j_{1}}\le 1$, $\Upsilon_{4}$ is a subset in index set $N$.
\item Relaxed proportional boundary:
\begin{equation}
\Gamma_{5}=\left\{\left(\begin{array}{cccc} {\overline{w}}_{1} & {\overline{w}}_{2} & \cdots & {\overline{w}}_{n} \end{array}\right)\in \Gamma _{0}|\frac{\overline{w}_{j_{1}}}{\overline{w}_{j_{2}}}+e_{(5)j_{1}j_{2}}^{-}\ge \delta_{j_{1}j_{2}}^{\prime\prime},j_{1}\in \Upsilon_{5}, j_{2}\in \Lambda _{5}\right\},\nonumber
\end{equation}
where $\delta_{j_{1}j_{2}}^{\prime\prime}$ is a constant and  $0\le \delta_{j_{1}j_{2}}^{\prime}\le 1$, $\Upsilon_{5}$ and $\Lambda _{5}$ are two disjoint subsets in index set $N$.
\end{enumerate}
And let $\Gamma$ be the sum of the five kinds of relaxed constraints: $\Gamma=\Gamma_{1}\cup\Gamma_{2}\cup\Gamma_{3}\cup\Gamma_{4}\cup\Gamma_{5}$.
}

\par{
Obviously, in the process of restricting attribute weights by experts, the less opinions experts have, the more favorable it is to determine the final attribute weights. In other words, we hope that these non-negative deviation variables are small enough. Combined with the above optimization model, a new optimization model can be established:
\begin{equation}
\begin{split}
&\max\left\{ CC(A_{1}), CC(A_{2}), \cdots, CC(A_{m})\right\}\\
&\min\left\{\sum_{j_{1},j_{2},j_{3}\in N}(e_{(1)j_{1}j_{2}}^{-}+e_{(2)j_{1}j_{2}}^{-}+ e_{(3)j_{1}j_{2}j_{3}}^{-}+ e_{(4)j_{1}}^{-}+ e_{(4)j_{1}}^{+}+ e_{(5)j_{1}j_{2}}^{-})\right\}\\
&s.t.
\begin{cases}
\left(\begin{array}{cccc} {\overline{w}}_{1} & {\overline{w}}_{2} & \cdots & {\overline{w}}_{n} \end{array}\right)\in \Gamma.\\
e_{(1)j_{1}j_{2}}^{-}\ge0,\quad j_{1}\in \Upsilon_{1},\quad j_{2}\in \Lambda _{1}.\\
e_{(2)j_{1}j_{2}}^{-}\ge0,\quad j_{1}\in \Upsilon_{2},\quad j_{2}\in \Lambda _{2}.\\
e_{(3)j_{1}j_{2}j_{3}}^{-}\ge0,\quad j_{1}\in \Upsilon_{3},\quad j_{2}\in \Lambda _{3},\quad j_{3}\in \Omega_{3}.\\
e_{(4)j_{1}}^{-}\ge0,\quad e_{(4)j_{1}}^{+}\ge0,\quad j_{1}\in \Upsilon_{4}.\\
e_{(5)j_{1}j_{2}}^{-}\ge0,\quad j_{1}\in \Upsilon_{5},\quad j_{2}\in \Lambda _{5}.
\end{cases}
\end{split}
\end{equation}
In order to facilitate the solution, the above model can be transformed into a single objective optimization model \cite{25}:
\begin{equation}\label{eq2}
\begin{split}
&\max\vartheta\\
&s.t.
\begin{cases}
CC(A_{i})\ge\vartheta,\quad i=1, 2, \cdots, m,\\
-\sum_{j_{1},j_{2},j_{3}\in N}(e_{(1)j_{1}j_{2}}^{-}+e_{(2)j_{1}j_{2}}^{-}+ e_{(3)j_{1}j_{2}j_{3}}^{-}+ e_{(4)j_{1}}^{-}+ e_{(4)j_{1}}^{+}+ e_{(5)j_{1}j_{2}}^{-})\ge\vartheta,\\
\left(\begin{array}{cccc} {\overline{w}}_{1} & {\overline{w}}_{2} & \cdots & {\overline{w}}_{n} \end{array}\right)\in \Gamma.\\
e_{(1)j_{1}j_{2}}^{-}\ge0,\quad j_{1}\in \Upsilon_{1},\quad j_{2}\in \Lambda _{1}.\\
e_{(2)j_{1}j_{2}}^{-}\ge0,\quad j_{1}\in \Upsilon_{2},\quad j_{2}\in \Lambda _{2}.\\
e_{(3)j_{1}j_{2}j_{3}}^{-}\ge0,\quad j_{1}\in \Upsilon_{3},\quad j_{2}\in \Lambda _{3},\quad j_{3}\in \Omega_{3}.\\
e_{(4)j_{1}}^{-}\ge0,\quad e_{(4)j_{1}}^{+}\ge0,\quad j_{1}\in \Upsilon_{4}.\\
e_{(5)j_{1}j_{2}}^{-}\ge0,\quad j_{1}\in \Upsilon_{5},\quad j_{2}\in \Lambda _{5}.
\end{cases}
\end{split}
\end{equation}
By combining subjective and objective methods to establish this optimization model, we not only consider the objective information contained in each index, but also consider the subjective opinions of experts. Therefore, the attribute weights with practical significance can be obtained to solve the multi-attribute group decision-making problem. At the same time, we can also give some other constraints to limit the attribute weight, such as the requirements of the decision-making project itself.
}

\subsection{The complete algorithm}
\begin{figure}[pos=h]
\centering
\includegraphics[height=7.7cm,width=12cm]{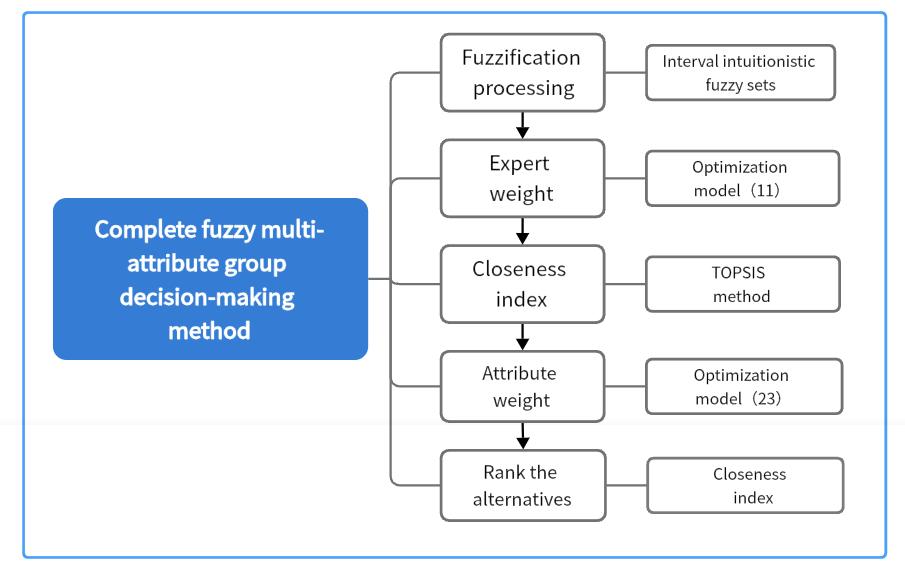}
\caption{Frame diagram}
\label{figure1}
\end{figure}

In this paper, interval intuitionistic fuzzy sets, optimization model for determining expert weights, improved TOPSIS decision-making method and optimization model for determining attribute weights are introduced respectively. This paper combines them for the first time, which can completely solve the multi-attribute group decision-making problem and help us make the final decision. The framework of this paper is shown in Figure \ref{figure1}.

\par{As can be seen from the figure: firstly, we use interval intuitionistic fuzzy sets to fuzzify the collected expert semantic evaluations; secondly, establish an optimization model to determine the weight of experts; then, the improved closeness index $CC(A_{i})$ is obtained by TOPSIS method; next, an optimization model is established by maximizing the closeness index $CC(A_{i})$ to determine the attribute weight; finally, the attribute weight is brought into the closeness index $CC(A_{i})$ so that the alternatives can be sorted. The complete algorithm is shown in algorithm \ref{algorithm}.}

\renewcommand{\algorithmicrequire}{\textbf{Input:}}
\renewcommand{\algorithmicensure}{\textbf{Output:}}
\begin{algorithm}
\caption{The complete algorithm}
\label{algorithm}
\begin{algorithmic}[1]
\Require Semantic evaluation of $n$ indicators of $m$ alternatives by $l$ experts.
\Ensure The ranking of $m$ alternatives.
\item The collected semantic evaluation is transformed into interval intuitionistic fuzzy set $A_{ij}^{k}(i=1,2,\cdots,m; j=1,2,\cdots,n; k=1,2,\cdots,l)$;
\item Calculate the distance $d_{(k)}=\left \| A^{(k)}-\boldsymbol{b} \right \|_{2}$, where $A^{(k)}$ is the evaluation result of all alternatives by the $k-$th expert $(k=1,2,\cdots,l)$ and $\boldsymbol{b}$ is the overall consistent score point;
\item Establish the optimization model (\ref{eq1}) and get the expert weight $\boldsymbol{w}=\left(\begin{array}{cccc} w_{1} &  w_{2} & \cdots &  w_{l}\end{array}\right)^{T}$;
\item Weighting the evaluation $A_{ij}^{k}$ to get ${\mathop{A}\limits _{\cdot}}_{ij}^{k}$, and calculating the optimal membership $p({\mathop{A}\limits _{\cdot}}_{ij}^{k})$ of ${\mathop{A}\limits _{\cdot}}_{ij}^{k}$;
\item The comprehensive decision matrix $D$ is obtained by using the extension IIOWA operator;
\item Find the positive and negative ideal solutions, and assume that the attribute weight is $\overline{\boldsymbol{w}}=\left(\begin{array}{cccc} \overline{w}_{1} &  \overline{w}_{2} & \cdots &  \overline{w}_{n}\end{array}\right)^{T}$ to get the closeness index $CC(A_{i})$ of the $i-$th alternative;
\item Establish the optimization model (\ref{eq2}) and get the attribute weight $\overline{\boldsymbol{w}}=\left(\begin{array}{cccc} \overline{w}_{1} &  \overline{w}_{2} & \cdots &  \overline{w}_{n}\end{array}\right)^{T}$;
\item According to the closeness index $CC(A_{i})$, $m$ alternatives can be ranked.
\end{algorithmic}
\end{algorithm}

\section{Case study}
This section will solve a multi-attribute group decision-making problem by using the method proposed above. We will use the same case as \cite{16} and compare the final results. This is a decision-making problem about the treatment of basilar artery occlusion in an 82-year-old solitary patient with hypertension. Her two sons and one daughter $\left\{D_{1}, D_{2}, D_{3}\right\}$ will make the linguistic evaluation of four alternatives $\left\{A_{1}, A_{2}, A_{3}, A_{4}\right\}$: intravenous thrombolysis, intra-arterial thrombolysis, antiplatelet therapy and heparinization from five indicators $\left\{x_{1}, x_{2}, x_{3}, x_{4}, x_{5}\right\}$: survival rate, severity of complications, possibility of cure, cost and possibility of recurrence, where $\left\{x_{1}, x_{3}\right\}$ are the benefit indicators and $\left\{x_{2}, x_{4}, x_{5}\right\}$ are the cost indicators, and the final decision will be obtained through the above algorithm. Table \ref{table1} shows the interval intuitionistic fuzzy sets corresponding to different linguistic evaluation.
\begin{table}[pos=h]
\begin{center}
\caption{Linguistic evaluation and corresponding interval intuitionistic fuzzy sets}
\label{table1}
\tabcolsep=2.1cm
\renewcommand\arraystretch{1.3}
\begin{tabular}{cc}
\hline
Linguistic evaluation \qquad & \qquad Interval intuitionistic fuzzy sets \\ 
\hline 
Very high(VH) \qquad&\qquad $<[0.75, 0.95], [0.00, 0.05]>$ \\
High(H) \qquad&\qquad $<[0.50, 0.70], [0.05, 0.25]>$  \\
Medium(M) \qquad&\qquad $<[0.30, 0.50], [0.20, 0.40]>$  \\
Low(L) \qquad&\qquad $<[0.05, 0.25], [0.50, 0.70]>$\\
Very low(VL) \qquad&\qquad $<[0.00, 0.05], [0.75, 0.95]>$\\
\hline
\end{tabular}
\end{center}
\end{table}

The algorithm proposed in this paper can be used to make the decision:
\begin{enumerate}[\bfseries{Step} 1.]
\item Collect the linguistic evaluation of four alternatives by three decision makers from five indicators, as shown in Table \ref{table2}, and convert them into interval intuitionistic fuzzy sets according to Table \ref{table1}.
\begin{table}[pos=h]
\begin{center}
\caption{Linguistic evaluation}
\label{table2}
\tabcolsep=1.2cm
\renewcommand\arraystretch{1.3}
\begin{tabular}{ccccc}
\hline
\multirow{2}{*}{Alternatives} & \multirow{2}{*}{Indicators} &\multicolumn{3}{c}{Decision makers}\\  
& & $D_{1}$  &  $D_{2}$  &  $D_{3}$\\ 
\hline 

\multirow{5}{*}{$A_{1}$}  &  $x_{1}$  &  VH  &  VH &  H\\

 & $x_{2}$   &  M  &  M &  L\\

& $x_{3}$  &  M  &  M &  H\\

 & $x_{4}$  &  M  &  M &  L\\

 & $x_{5}$  &  M  &  L &  M\\
\hline

\multirow{5}{*}{$A_{2}$}  &  $x_{1}$  &  H  &  H &  VH\\

 & $x_{2}$   &  M  &  H &  M\\

& $x_{3}$  &  VH  & H &  VH\\

 & $x_{4}$  &  VH  &  VH &  H\\

 & $x_{5}$  &  L  &  L &  VL\\
\hline

\multirow{5}{*}{$A_{3}$}  &  $x_{1}$  &  M  &  L &  M\\

 & $x_{2}$   &  L  &  L &  M\\

& $x_{3}$  &  VL  &  VL &  L\\

 & $x_{4}$  &  L  &  VL &  VL\\

 & $x_{5}$  &  H  &  VH &  VH\\
\hline

\multirow{5}{*}{$A_{4}$}  &  $x_{1}$  &  M  &  H &  H\\

 & $x_{2}$   &  VL  &  M &  L\\

& $x_{3}$  &L  &  M &  L\\

 & $x_{4}$  &  M  &  H &  L\\

 & $x_{5}$  &  VH  &  H &  VH\\
\hline

\multirow{5}{*}{$A_{5}$}  &  $x_{1}$  &  VH  &  VH &  H\\

 & $x_{2}$   &  M  &  M &  L\\

& $x_{3}$  &  M  &  M &  H\\

 & $x_{4}$  &  M  &  M &  L\\

 & $x_{5}$  &  M  &  L &  M\\
\hline
\end{tabular}
\end{center}
\end{table}

\item By splitting the interval intuitionistic fuzzy sets, the length of the evaluation column vector of the $k-$th decision-maker for the $i-$th alternative will be four times as long as the original one. And calculate the consistent score point $\boldsymbol{b}_{(i)}$ corresponding to the $i-$th alternative from the linear combination of $l$ experts' evaluations. Then, assemble the evaluation results of all alternatives and calculate the distance from the evaluation results $A^{k}$ of the $k-$th expert to the overall consistent score point $\boldsymbol{b}$.
\item Establish the expert weight optimization model:
\begin{equation}\nonumber
\begin{split}
&\min_{\boldsymbol{w}} Q(\boldsymbol{w})=\sum_{k=1}^{3}d_{(k)}\\
&s.t.\begin{cases}
\sum_{k=1}^{3}w_{k}=1,\\
0\le w_{k}\le 1, \quad 1\le k \le 3.
\end{cases}
\end{split}
\end{equation}
It can be obtained that the weight of three decision makers is $\boldsymbol{w}=\left(\begin{array}{ccc} 0.45456 &  0.26647 &  0.27897\end{array}\right)^{T}$, which is different from that given directly as $\left(\begin{array}{ccc} 0.40 &  0.35 &  0.25\end{array}\right)^{T}$ in \cite{16}. Table \ref{table3} shows the weights of the three decision makers and the distance to the overall consistent score point. Through observation, it is found that the greater the weight of the decision maker, the closer his evaluation is to the overall consistent score point, which is in line with our goal of establishing the optimization model (\ref{eq1}).
\begin{table}[pos=h]
\begin{center}
\caption{Expert weight and distance}
\label{table3}
\tabcolsep=0.8cm
\renewcommand\arraystretch{1.3}
\begin{tabular}{|c|c|c|c|}
\hline
Decision maker & $D_{1}$  & $D_{2}$  &  $D_{3}$\\ 
\hline 
Weight  &  0.45456  &   0.26647  &   0.27897\\
\hline
Distance to overall consistent score point $\boldsymbol{b}$  &  0.71326  &  1.21682  &  1.16203\\
\hline
\end{tabular}
\end{center}
\end{table}
\item Through the expert weights obtained above, the evaluation $A_{ij}^{k}$  is weighted by (\ref{eq3}) to get ${\mathop{A}\limits _{\cdot}}_{ij}^{k}$ , and the optimal membership $p({\mathop{A}\limits _{\cdot}}_{ij}^{k})$ of ${\mathop{A}\limits _{\cdot}}_{ij}^{k}$ is calculated by (\ref{eq4}). For example, ${\mathop{A}\limits _{\cdot}}_{11}^{1}=<[0.8490, 0.9832], [0, 0.0168]>$, $p({\mathop{A}\limits _{\cdot}}_{11}^{1}\supseteq {\mathop{A}\limits _{\cdot}}_{11}^{2})=0.6650$, $p({\mathop{A}\limits _{\cdot}}_{11}^{1}\supseteq {\mathop{A}\limits _{\cdot}}_{11}^{3})=0.9168$ and $p({\mathop{A}\limits _{\cdot}}_{11}^{1})=0.4303$.
\item A comprehensive decision matrix $D$ (whose weighted vector is $\tau=\left(\begin{array}{ccc} 0.2429 &  0.5142 &  0.2429\end{array}\right)^{T} $) is obtained by using the extension IIOWA:
\begin{equation}
D=\begin{pmatrix}
 <[0.6896, 0.9153], [0, 0.0816]> & \cdots  &<[0.2454, 0.4422], [0.2565, 0.4644]> \\
  \vdots & \ddots  & \vdots \\
 <[0.4088, 0.6189], [0.0983, 0.3668]>& \cdots  &<[0.6959, 0.9192], [0, 0.0780]>
\end{pmatrix}. \nonumber
\end{equation}
\item Use (\ref{eq5.1}) and (\ref{eq5.2}) to find the positive and negative ideal solutions of interval intuitionistic fuzzy corresponding to the comprehensive decision matrix $D$, and calculate the closeness of each alternative through (\ref{eq6}).
\item \label{s7}Establish the attribute weight optimization model:
\begin{equation}
\begin{split}
&\max\vartheta\\
&s.t.
\begin{cases}
\frac{0.9492\overline{w}_{1}+0.6887\overline{w}_{2}+\overline{w}_{3}+0.9553\overline{w}_{4}+0.9549\overline{w}_{5}}{1.4492\overline{w}_{1}+1.4718\overline{w}_{2}+1.8308\overline{w}_{3}+1.8609\overline{w}_{4}+1.7735\overline{w}_{5}}\ge\vartheta,
\\[5pt]
\frac{0.8645\overline{w}_{1}+0.5\overline{w}_{2}+\overline{w}_{3}+0.5\overline{w}_{4}+\overline{w}_{5}}{1.4894\overline{w}_{1}+1.4165\overline{w}_{2}+1.5\overline{w}_{3}+1.5\overline{w}_{4}+1.5\overline{w}_{5}}\ge\vartheta,
\\[5pt]
\frac{0.5\overline{w}_{1}+0.7906\overline{w}_{2}+0.5\overline{w}_{3}+\overline{w}_{4}+0.5429\overline{w}_{5}}{1.4492\overline{w}_{1}+1.4450\overline{w}_{2}+1.5\overline{w}_{3}+1.5\overline{w}_{4}+1.5429\overline{w}_{5}}\ge\vartheta,
\\[5pt]
\frac{0.6817\overline{w}_{1}+0.8817\overline{w}_{2}+0.7866\overline{w}_{3}+0.8495\overline{w}_{4}+0.5\overline{w}_{5}}{1.4745\overline{w}_{1}+1.4442\overline{w}_{2}+1.7866\overline{w}_{3}+1.8495\overline{w}_{4}+1.5\overline{w}_{5}}\ge\vartheta,
\\[5pt]
(e_{(i)14}^{-}+e_{(ii)52}^{-}+ e_{(iii)324}^{-}+ e_{(iv)4}^{-}+ e_{(iv)4}^{+}+ e_{(v)23}^{-})\ge\vartheta,\\
\overline{w}_{1}+e_{(i)14}^{-}\ge\overline{w}_{4},\quad
\overline{w}_{5}-\overline{w}_{2}+e_{(ii)52}^{-}\ge0.04,\quad
\overline{w}_{3}-2\overline{w}_{2}+\overline{w}_{4}+e_{(iii)324}^{-}\ge0,\\
\overline{w}_{4}+e_{(iv)4}^{-}\ge0.08,\quad
\overline{w}_{4}-e_{(iv)4}^{+}\le0.15,\quad
\frac{\overline{w}_{2}}{\overline{w}_{3}}+e_{(v)23}^{-}\ge0.4,\\
(e_{(i)14}^{-}\ge0,\quad e_{(ii)52}^{-}\ge0,\quad  e_{(iii)324}^{-}\ge0,\quad  e_{(iv)4}^{-}\ge0,\quad e_{(iv)4}^{+}\ge0,\quad e_{(v)23}^{-})\ge0,\\
\overline{w}_{1}+\overline{w}_{2}+\overline{w}_{3}+\overline{w}_{4}+\overline{w}_{5}=1,\\
\overline{w}_{j}\ge0,\quad j=1,2, \cdots, 5.
\end{cases}
\end{split}\nonumber
\end{equation}
Then the weight of five attributes is $\overline{\boldsymbol{w}}=\left(\begin{array}{ccccc} 0.2234 &  0.1659 &  0.2245 &0.1074&0.2787\end{array}\right)^{T}$. 

\item Substitute the attribute weights obtained in step \ref{s7} into the closeness index to obtain: $CC(A_{1})=0.5575228$, $CC(A_{2})=0.5686608$, $CC(A_{3})=0.4058395$, $CC(A_{4})=0.4583039$. According to the closeness, the ranking of each alternative is $A_{2}>A_{1}>A_{4}>A_{3}$, thus the optimal alternative is $A_{2}$.
\end{enumerate}

\par{
Compared with \cite{16}, the improvement of this paper is combined with \cite{21}. The expert weight obtained by establishing the optimization model through subjective and objective weighting method is more convincing than giving the expert weight directly through subjective weighting method. And from the same case, the closeness obtained in \cite{16} is: $A_{2}>A_{1}>A_{4}>A_{3}$. Although it is slightly different from the closeness calculated in this paper, the final ranking of the closeness of each alternative is consistent, and the optimal alternative is $A_{2}$.
}

\section{Conclusion}
In the context of interval-valued intuitionistic fuzzy sets, this paper extends the optimization model for determining expert weights proposed in \cite{21} and combines it with the method proposed in \cite{16}. In this way, the determination of expert weight and attribute weight gives full play to the advantages of subjective and objective weighting methods. By combining with TOPSIS, a complete fuzzy multi-attribute group decision-making method is formed. The feasibility of the method proposed in this paper is verified by calculating the decision-making problem of treatment scheme in \cite{16}. And compared the results of this paper with those of \cite{16}, it is found that the final decision-making results are consistent, which verifies the effectiveness of the proposed method of this paper. Our next research direction is to continue to improve the optimization model for determining expert weights proposed in \cite{21}, so as to be used in other multi-attribute group decision-making methods to improve the decision-making effect.

\section*{Acknowledgements}
This work is supported by the ``Key Research Program'' (Grant No.2022YFC3801300), the Ministry  of Science and Technology, PRC. And we would like to deliver thanks to Professor Han Huilei and Professor Huang Li for their assistance.

\bibliographystyle{unsrt}

\bibliography{cas-refs}



\end{document}